\documentclass[review]{elsarticle}
\usepackage{threeparttable}
\usepackage{longtable}
\usepackage{adjustbox}
\usepackage{multirow}
\usepackage{hyperref}
\newcommand{\abbreviations}[1]{%
\nonumnote{\textit{Abbreviations:\enspace}#1}}

\journal{Lung Cancer}
\begin{document}

\begin{frontmatter}

\title{Relationship between pulmonary nodule malignancy and surrounding pleurae, airways and vessels: a quantitative study using the public LIDC-IDRI dataset}

\author[mymainaddress,mysecondaryaddress,mythirdaddress]{Yulei Qin}
\cortext[mycorrespondingauthor]{Corresponding author}
\ead{qinyulei@sjtu.edu.cn}
%\corref{mycorrespondingauthor}

\author[mymainaddress,mysecondaryaddress]{Yun Gu}
\ead{geron762@sjtu.edu.cn}

\author[mysecondaryaddress]{Hanxiao Zhang}
\ead{hanxiao.zhang@sjtu.edu.cn}

\author[mymainaddress,mysecondaryaddress]{Jie Yang\corref{mycorrespondingauthor}}
\ead{jieyang@sjtu.edu.cn}

\author[lihuiwangaddress]{Lihui Wang}
\ead{wlh1984@gmail.com}

\author[fengyaoaddress]{Zhexin Wang}
\ead{wangzhexin001@hotmail.com}

\author[fengyaoaddress,mysecondaryaddress]{Feng Yao}
\ead{yaofeng6796678@126.com}

\author[mythirdaddress]{Yue-Min Zhu}
\ead{zhu@creatis.insa-lyon.fr}

\address[mymainaddress]{Institute of Image Processing and Pattern Recognition, Shanghai Jiao Tong University, Shanghai 200240, China}
\address[mysecondaryaddress]{Institute of Medical Robotics, Shanghai Jiao Tong University, Shanghai 200240, China}
\address[mythirdaddress]{CREATIS, INSA Lyon, CNRS UMR 5220, INSERM U1206, Universit\'{e} de Lyon, Villeurbanne 69621, France}
\address[lihuiwangaddress]{Key Laboratory of Intelligent Medical Image Analysis and Precise Diagnosis of Guizhou Province, School of Computer Science and Technology, Guizhou University, Guiyang 550025, China}
\address[fengyaoaddress]{Department of Thoracic Surgery, Shanghai Chest Hospital, Shanghai Jiao Tong University, Shanghai 200025, China}

\begin{abstract}

\textbf{Objectives}: To investigate whether the pleurae, airways and vessels surrounding a nodule on non-contrast computed tomography (CT) can discriminate benign and malignant pulmonary nodules. 

\textbf{Materials and Methods}: The LIDC-IDRI dataset, one of the largest publicly available CT database, was exploited for study. A total of 1556 nodules from 694 patients were involved in statistical analysis, where nodules with average scorings $<$3 and $>$3 were respectively denoted as benign and malignant. Besides, 339 nodules from 113 patients with diagnosis ground-truth were independently evaluated. Computer algorithms were developed to segment pulmonary structures and quantify the distances to pleural surface, airways and vessels, as well as the counting number and normalized volume of airways and vessels near a nodule. Odds ratio (OR) and Chi-square ($\chi^2$) testing were performed to assess the correlation between features of surrounding structures and nodule malignancy. A non-parametric receiver operating characteristic (ROC) analysis was conducted in logistic regression to evaluate discrimination ability of each structure.

\textbf{Results}: For the benign and malignant groups, the average distances from nodules to pleural surface, airways and vessels are respectively (6.56, 5.19), (37.08, 26.43) and (1.42, 1.07) mm. The correlation between nodules and the counting number of airways and vessels that contact or project towards nodules are respectively (OR=22.96, $\chi^2$=105.04) and (OR=7.06, $\chi^2$=290.11). The correlation between nodules and the volume of airways and vessels are (OR=9.19, $\chi^2$=159.02) and (OR=2.29, $\chi^2$=55.89). The areas-under-curves (AUCs) for pleurae, airways and vessels are respectively 0.5202, 0.6943 and 0.6529.

\textbf{Conclusion}: Our results show that malignant nodules are often surrounded by more pulmonary structures compared with benign ones, suggesting that features of these structures could be viewed as lung cancer biomarkers.

\end{abstract}


\abbreviations{CT: Computed tomography; LIDC: Lung image database consortium; IDRI: Image database resource initiative; XML: Extensible markup language; 2-D: two-dimensional; 3-D: three-dimensional.}

\end{frontmatter}

\section{Introduction}

% basic background on lung cancer
The latest 2020 Global Cancer Statistics demonstrate that lung cancer remains the leading cause of cancer death worldwide \cite{sung2021global}. The outcomes of lung cancer are highly dependent on the stage. Although the 5-year survival rate is 6\% for patients with metastatic disease, if early diagnosis and treatment are made, the survival rate could be greatly increased to 60\% \cite{wu2016lung, siegel2021cancer}. The National Lung Screening Trial shows that annual screening with computed tomography (CT) brings about a 20\% reduction in lung cancer mortality \cite{national2011reduced}.

%It accounts for approximately 11.4\% of 19.3 million new cancer cases and 18.0\% of new deaths related to cancer. 
%for localized stage disease
% for both men and women

% basic literature about pulmonary nodule classification
Pulmonary nodule appears as a white spot inside lung on chest CT \cite{american2016lung}. Its likelihood of malignancy indicates lung cancer. To reduce the burden of radiologists in reading and assessing nodules on a slice-by-slice basis, many researchers have proposed computer algorithms for nodule malignancy classification \cite{mori1990small, gurney1995solitary, mcnitt1999pattern, nakamura2000computerized, matsuki2002usefulness, matsuoka2005peripheral, suzuki2005computer, okada2009computer, chen2010neural, emaminejad2015fusion, shen2015multi, kirby2016lungx, farag2017feature, nibali2017pulmonary, shaukat2017fully, xie2018knowledge, xie2019semi, zhang2020learning}. Most of these methods relied on handcrafted features such as nodule intensity, sphericity and texture descriptors. Based on extracted features, classifiers like support vector machine and logistic regression were employed for malignancy estimation. Recently, deep learning methods have prevailed in the nodule classification task, where features are learned automatically from CT images. Such superior performance comes with a side effect of relatively low interpretability.

% problem
In the development of automatic classification methods, false positive predictions cause harmful consequences and unnecessary treatments (e.g., follow-up CT scans and invasive biopsies) \cite{park2013examining, thalanayar2015indolent}. To improve performance, one potential solution is to incorporate clinically relevant context information as much as possible into nodule malignancy estimation. Most current methods only refer to the size, growth rate and morphologic characteristics of nodules since these attributes have been proved effective in diagnosis \cite{larici2017lung}. However, the context of a nodule, namely the relationship between a nodule and its surrounding structures, might provide supplementary diagnostic basis and should not be neglected. The exploration and understanding of such relationship is of high priority for both medical and computer vision researchers.
%s, which represent non-cancerous nodules but diagnosed as malignant, 

\subsection{Related work}

\paragraph{Relationship between a nodule and pleurae} Kim et al. \cite{kim2019clinical} evaluated whether the attachment of a nodule to pleura related to visceral pleural invasion (VPI). The increase of solid portion in part-solid nodules suggested high risk of VPI. Heidinger et al. \cite{heidinger2019visceral} showed that solid nodules that contacted pleural surface were associated with higher likelihood of VPI than part-solid nodules. Zhu et al. \cite{zhu2020management} performed a retrospective review and found that all non-calcified solid nodules attached to the costal pleura were benign if they had smooth margins,  oval, semi-circular or triangular shapes, with diameters less than 10 mm.

\paragraph{Relationship between a nodule and airways} Gaeta et al. \cite{gaeta1991bronchus} investigated the value of bronchus sign in predicting the success of transbronchial biopsy and brushing for patients with peripheral lung lesions. They later studied how the type of tumor-bronchus relationship will determine the yield of transbronchial biopsy \cite{gaeta1993carcinomatous}. Qiang et al. \cite{qiang2004relationship} used multi-slice CT to study five different tumor-bronchus relationships. The categorization of such relationship reflected the pathological changes of nodules to some extent. Cui et al. \cite{cui2009value} defined four types of nodule-bronchus relationship by bronchial morphology and confirmed their roles in determining the degree of differentiation of solid pulmonary nodules.

%in position and shape

\paragraph{Relationship between a nodule and vessels} Mori et al. \cite{mori1990small} studied tumor angiogenesis and reported that the involvement of veins converging to a nodule was strongly suggestive of malignancy. Kawata et al. \cite{kawata2000differential} used geometry-based vector fields for quantification instead of explicitly segmenting vascular structures around nodules. Wang et al. \cite{wang2011relationship} demonstrated that the morphology subtypes of bronchi, pulmonary arteries and veins correlated with size, location, pathology and stage of peripheral lung cancer. Wang et al. \cite{wang2017vasculature} validated that the vessels surrounding a nodule could help discriminate between benign and malignant nodules. The position and orientation of vessels relative to a nodule may imply lung cancer stage and pathology \cite{yang1999small, rigau2002blood, gao2013multi}.

%proposed a scheme to 
%Reports in literature also showed that 
%analyzed nodule surroundings. They 

\subsection{Challenges} There exist limitations in previous studies on the relationship between the malignancy of a nodule and its surrounding airways, vessels and pleurae. First, the size and diversity of CT database are limited in early work. Due to the difficulty of data collection and privacy issues, the number of patient cases is usually less than 100 and multi-center study was not available. Second, the extraction of pulmonary structures is a challenge without the help of image segmentation algorithms. It is tedious and time-consuming for radiologists to manually delineate and measure airways, vessels and pleurae from CT, which in turn imposes restrictions on the size of database. Third, the relationship between a nodule and other structures is often categorized into subtypes by descriptive definition. Very few studies computed features to quantify such relationship. Without quantitative measurement, it is difficult to discover the value of surrounding structures in assessing nodule malignancy. Last but not least, the correlation analyses on categorical or dichotomized variables may not be adequate to reveal the discrimination ability of airways, vessels and pleurae. Classifiers are needed to evaluate the accuracy of utilizing such relationship for benign-malignant classification.

\subsection{Contributions}

In this study, we investigated if a pulmonary nodule's surrounding structures like pleurae, airways and vessels (arteries and veins) could discriminate between benign and malignant nodules. Deep learning methods were developed to segment lung field (pleural surface), tubular airways and vessels (arteries and veins) from 3-D CT scans. Given the position and diameter of each nodule, a local volume-of-interest (VOI) was extracted to quantify features of pulmonary structures inside. Then, statistical correlation analyses and logistic regression experiments were performed to evaluate whether the presence of certain surrounding structures will be more frequent with malignant nodules than with benign ones. The database involved is one of the largest publicly available LIDC-IDRI lung nodule dataset \cite{armato2011lung}. According to our study on the database demographics and annotations, extensive experiments were designed and conducted. It is believed that investigation of such relationship on a large-scale database, by means of quantitative analysis, is indispensable and enlightening for improving the accuracy of nodule diagnosis.

\section{Materials and Methods}

\subsection{Data sources}

The LIDC-IDRI database \cite{armato2011lung} is one of the largest publicly available lung nodule dataset, consisting of 1018 CT scans collected retrospectively from seven medical centers. These images were acquired by a wide range of scanner manufacturers and reconstruction parameters \cite{gupta2018automatic}, with slice thickness ranging from 0.6 to 5.0 mm. The size of axial slices of all CT scans is 512$\times$512. 

%It is diverse in terms of image acquisition and reconstruction parameters and nodule candidates . 
%This study evaluated pulmonary nodules on chest CT scans from LIDC-IDRI dataset \cite{armato2011lung}. 
%(e.g., scanner manufacturer, slice thickness, dose level, tube voltage and current)
%, nodule type and patient health condition

\subsection{Annotations and diagnosis results}

% In order to reflect the inter-observer variability, 

A two-phase annotation process was adopted. In the first phase, each radiologist independently marked suspicious lesions on CT as non-nodule, nodule (diameter$<$3 mm) and nodule (diameter$\geq$3 mm). For nodules (diameter$\geq$3 mm), their boundaries were outlined and nine attributes (subtlety, internal structure, calcification, sphericity, margin, lobulation, spiculation, texture and malignancy) were subjectively estimated as scorings. For nodules (diameter$<$3 mm), only the coordinates of their centroids were provided. In the second phase, each radiologist reviewed annotations with reference to anonymized annotations from other radiologists. Modifications were allowed but no consensus among observers was enforced. All annotations were encoded in XML files.

%read CT images and 
%The LIDC-IDRI organizers adopted 
%(unblind reading)
%(blind reading)
%on marks and scorings
%Apart from annotations from radiologists, 

We also selected a subset of 113 patient cases with diagnosis-definite ground-truth labels from LIDC-IDRI. The patient-level diagnostic results include: 1) benign; 2) malignant, primary lung cancer; 3) malignant metastatic. Diagnosis methods involve: 1) review of radiological images to show 2 years of stable nodule; 2) biopsy; 3) surgical resection; 4) progression or response.

% or non-malignant
%The LIDC-IDRI also provides  of 113 . 

\subsection{Data processing and experiment design}

Statistical analysis and logistic regression experiments were respectively designed with different ways of dataset partition and processing.

%Considering that LIDC-IDRI dataset contains both subjective annotations and objective definite diagnosis labels, 

\paragraph{Statistical analysis} Following the procedures in previous studies \cite{han2015texture, dhara2016combination, hussein2017tumornet, shen2017multi, shen2016learning, xie2018knowledge, xie2019semi}, all nodules (diameter$\geq$3 mm) with at least one annotation were extracted from XML files and isometric resampling of $1.0\times1.0\times1.0$ mm$^3$ was performed on CT scans to tackle a large variety of spatial resolution. The malignancy of each nodule was scored from 1 (highly unlikely malignant) to 5 (highly suspicious) and scorings from different radiologists were averaged. Nodules with mean malignancy score $<$3, $=$3 and $>$3 are respectively benign, uncertain and malignant. In total, there are 2505 nodules (1044 benign, 512 malignant and 949 uncertain). Note that the number of nodules extracted from the LIDC-IDRI database may vary from study to study (e.g., 1356 \cite{han2015texture}, 891 \cite{dhara2016combination}, 1145 \cite{hussein2017tumornet}, 2618 \cite{shen2017multi}, 2272 \cite{shen2016learning}, 2557 \cite{xie2018knowledge}, 2669 \cite{armato2011lung}). The reasons behind are threefold: 1) the database was released in 2011 and has been updated in 2012, 2015, 2018 and 2020; 2) different software packages were developed for analyzing XML files; 3) various strategies of label fusion were adopted without establishing any single criterion. All these interpretations of data formatting were possible and justified \cite{fedorov2019standardized, fedorov2020dicom}. To reduce the impact of nodules with uncertain malignancy, we excluded these uncertain nodules in experiments. All the remaining 1556 nodules from 694 patients were involved in statistical analysis.

%or enforcing
%and the current processing step is justified. XML
%extended and
%the malignancy  for each nodule. 
%over five levels, 
%In view of the diversity of CT images,  . The slice thickness and voxel spacing were normalized to .
%To discover the relationship between surrounding structures and nodule malignancy, 
%due to different packages for processing XML files, different strategies for label fusion and different criteria on image quality, 
%via trilinear interpolation

Although the malignancy scorings are subjective, they do statistically reflect certain characteristics of nodules with malignant tendency. Radiologists comprehensively considered multiple nodule attributes to assess malignancy. It is reasonable to refer to such scorings as \textbf{proxy} ground-truth labels. Besides, to quantify inter-observer variability, we followed Ref. \cite{horsthemke2009evaluation, lin2017measuring} to calculate the mean rating difference of each nodule between all pairs of radiologists. For example, given scorings \{1,2,4\} from three observers, the paired differences are \{$\mid$1-2$\mid$, $\mid$1-4$\mid$, $\mid$2-4$\mid$\} and the mean value is 2. The distribution of nodules with proxy malignancy label is summarized in Table \ref{table:distribution1}.

%calculate the mean absolute difference 
%among scorings of each nodule. 
%absolute
%in the present study
% of malignancy estimation

\begin{table}[htbp]
\centering
\caption{Subject demographics in statistical analysis experiments}
\label{table:distribution1}
\begin{adjustbox}{width=\columnwidth}
\begin{threeparttable}
\begin{tabular}{lllll}
\hline
Nodule & \begin{tabular}[c]{@{}l@{}}Benign\\ scoring$<$3\end{tabular} & \begin{tabular}[c]{@{}l@{}}Malignant\\ scoring$>$3\end{tabular} & \begin{tabular}[c]{@{}l@{}}Uncertain\\ scoring=3\end{tabular} & All \\ \hline
No. &  1044  &  512  &  949  &  2505 \\
Equivalent diameter (Eqd.)\tnote{*} &  &  &  &  \\
\quad Avg. eqd.\tnote{$\dagger$} &   6.97$\pm$2.93 & 14.52$\pm$6.47  &  7.50$\pm$3.20  & 8.64$\pm$5.00 \\
\quad No. eqd.$\leq$10 mm &  965  &  143  &  808   &  1916 \\
\quad No. 10$<$eqd.$\leq$20 mm &  71  &  265  & 130   &  466 \\
\quad No. eqd.$>$20 mm &  8 &  104  & 11  & 123  \\
Texture\tnote{$\dagger$} &  &  &  &  \\
\quad Solid &  766  &  324 &  568  &  1658  \\
\quad Part-solid &  278   &  188   &  381   &   847  \\
Scorings &  &  &  &  \\
\quad No. radiologists =1 & 316   &   66   &   284   &  666  \\
\quad No. radiologists $>$1 & 728  &  446  &  665  &  1839 \\
\quad Avg. difference\tnote{$\dagger$} &  0.78$\pm$0.63  &  1.05$\pm$0.54 &  0.77$\pm$0.59  &  0.84$\pm$0.61 \\ 
Background lung diseases &  &  &  &  \\
\quad Emphysema & 93 & 40 & 79 & 212 \\
\quad Fibrosis & 32 & 19 & 27 & 78 \\
%\quad Bronchial asthma & 1 & 6 & 1 & 8 \\
\quad Pulmonary congestion & 103 & 83 & 92 & 278 \\
\hline
\end{tabular}
\begin{tablenotes}
\item[*]\footnotesize Given the volume of a nodule, equivalent diameter (Eq. diameter, Eqd.) is calculated by assuming the nodule as an ideal sphere. Compared with nodule's maximum diameter, the equivalent diameter is a normalized metric and reflects more closely its actual size.
\item[$\dagger$]\footnotesize Statistical significance (p-value$<$0.05) was observed between different nodule groups.
\end{tablenotes}
\end{threeparttable}
\end{adjustbox}
\end{table}
%\begin{tabular}[c]{@{}l@{}}  \end{tabular}

\begin{table}[htbp]
\centering
\caption{Subject demographics in logistic regression experiments}
\label{table:distribution2}
%\begin{adjustbox}{width=\columnwidth}
\begin{threeparttable}
\begin{tabular}{llll}
\hline
Nodule & \begin{tabular}[c]{@{}l@{}}Benign\\ patient-level\end{tabular} & \begin{tabular}[c]{@{}l@{}}Malignant\\ patient-level\end{tabular}  & All \\ \hline
No. &  81  &  258   &  339 \\
Equivalent diameter (Eqd.)  &  &  &  \\
\quad Avg. eqd.\tnote{$\dagger$}  &   7.93$\pm$3.83  &   10.57$\pm$5.90  &  9.94$\pm$5.59   \\
\quad No. eqd.$\leq$10 mm &  67  &  160  &  227 \\
\quad No. 10$<$eqd.$\leq$20 mm &  12  &  74   &  86 \\
\quad No. eqd.$>$20 mm &  2   &   24  &  26  \\
Texture &  &     &  \\
\quad Solid &  69   &  208    &  277   \\
\quad Part-solid &  12   &  50     &   62   \\
Scorings &  &     &  \\
\quad No. radiologists =1 & 23   &   48     &  71  \\
\quad No. radiologists $>$1 & 58   &  210     &  268 \\
\quad Avg. difference\tnote{$\dagger$}  &  0.75$\pm$0.60   &  0.90$\pm$0.58    &  0.87$\pm$0.59  \\ 
Background lung diseases &  &  &  \\
\quad Emphysema & 8 & 22 & 30  \\
\quad Fibrosis & 2 & 12 & 14  \\
%\quad Bronchial asthma & 0 & 0 & 0  \\
\quad Pulmonary congestion & 6 & 46 & 52  \\
\hline
\end{tabular}
\begin{tablenotes}
\item[$\dagger$]\footnotesize Statistical significance (p-value$<$0.05) was observed between different nodule groups.
\end{tablenotes}
\end{threeparttable}
%\end{adjustbox}
\end{table}

\paragraph{Logistic regression} To verify if the surrounding structures can predict the malignancy of a nodule, a logistic regression classifier \cite{kleinbaum2002logistic} was developed for receiver-operating-characteristic (ROC) analysis. We used 113 patient cases with diagnosis results as the testing set. Patients with diagnosis category=1 are benign and those with category=2 or 3 are malignant. Note that diagnosis was performed at \textbf{patient-level} as a multiple-instance learning (MIL) task \cite{shen2016learning}. For a patient with multiple nodule instances, he/she is malignant if at least one instance is malignant. Only if all instances are benign, the patient is benign. During training, we used 914 benign nodules and 429 malignant nodules. The classifier learned to predict the probability of each nodule being malignant. During testing, for each patient, the maximum of malignancy probabilities of all visible nodules was deemed as patient-level malignancy probability. The training set (1343 nodules) does not overlap with testing set (339 nodules). The distribution of nodules with patient-level ground-truth is summarized in Table \ref{table:distribution2}.

%pleurae, airways and vessels
%In the light of this statement,

%\begin{tabular}[c]{@{}l@{}}  absolute\\ \end{tabular} 

\subsection{Quantification of surrounding structures}

\begin{figure}[htbp]
\centering
\includegraphics[width=\columnwidth]{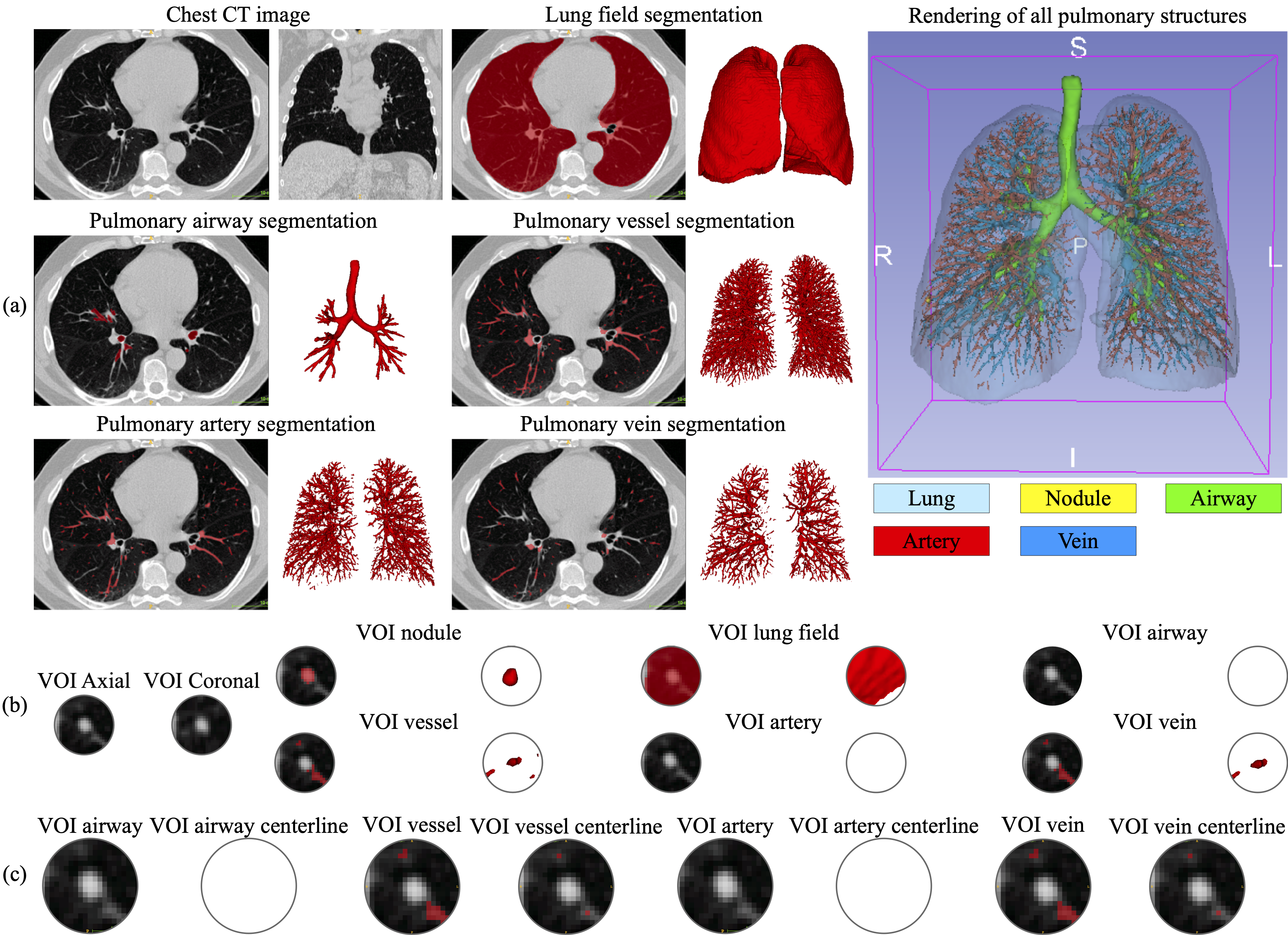}
\caption{Extraction of lung fields, airways and vessels. (a) Masks of pulmonary structures were generated via segmentation methods. (b) A VOI was cropped based on the nodule position and diameter. (c) The centerlines of airways and vessels were computed. The blank region inside VOI means there exist no target structures. The segmented airways and vessels include both lumen and wall.}
\label{fig:segmentation}
\end{figure}

%maximum

The lung fields, airways and vessels (arteries and veins) were segmented on CT scans using deep learning methods \cite{qin2020learning, qin2020learning2}. The centerlines of airway and vessel branches were extracted via skeletonization \cite{lee1994building}. The coordinates and masks of nodules were obtained from LIDC-IDRI annotations. For each nodule, a spherical sub-volume located at the nodule centroid was cropped as VOI. Its radius is the nodule diameter and has to be at least 6 mm. Fig. \ref{fig:segmentation} illustrates the steps of extracting pulmonary structures. To describe the proximity between a nodule and its surrounding structures, we measured the distance from the nodule surface to its closest target structure.

%maximum
%For each CT scan, 
%Compared with the distance from nodule centroid, the distance from surface avoids the impact of nodule size on measurement.
%as previously described
%After segmentation, t
%manual pulmonary
%from CT volume 
%of the VOI
%calculated the minimum
%If nodules are attached to certain structure, the corresponding distance is 1 mm.
% from the nodule VOI

\begin{figure}[htbp]
\centering
\includegraphics[width=\columnwidth]{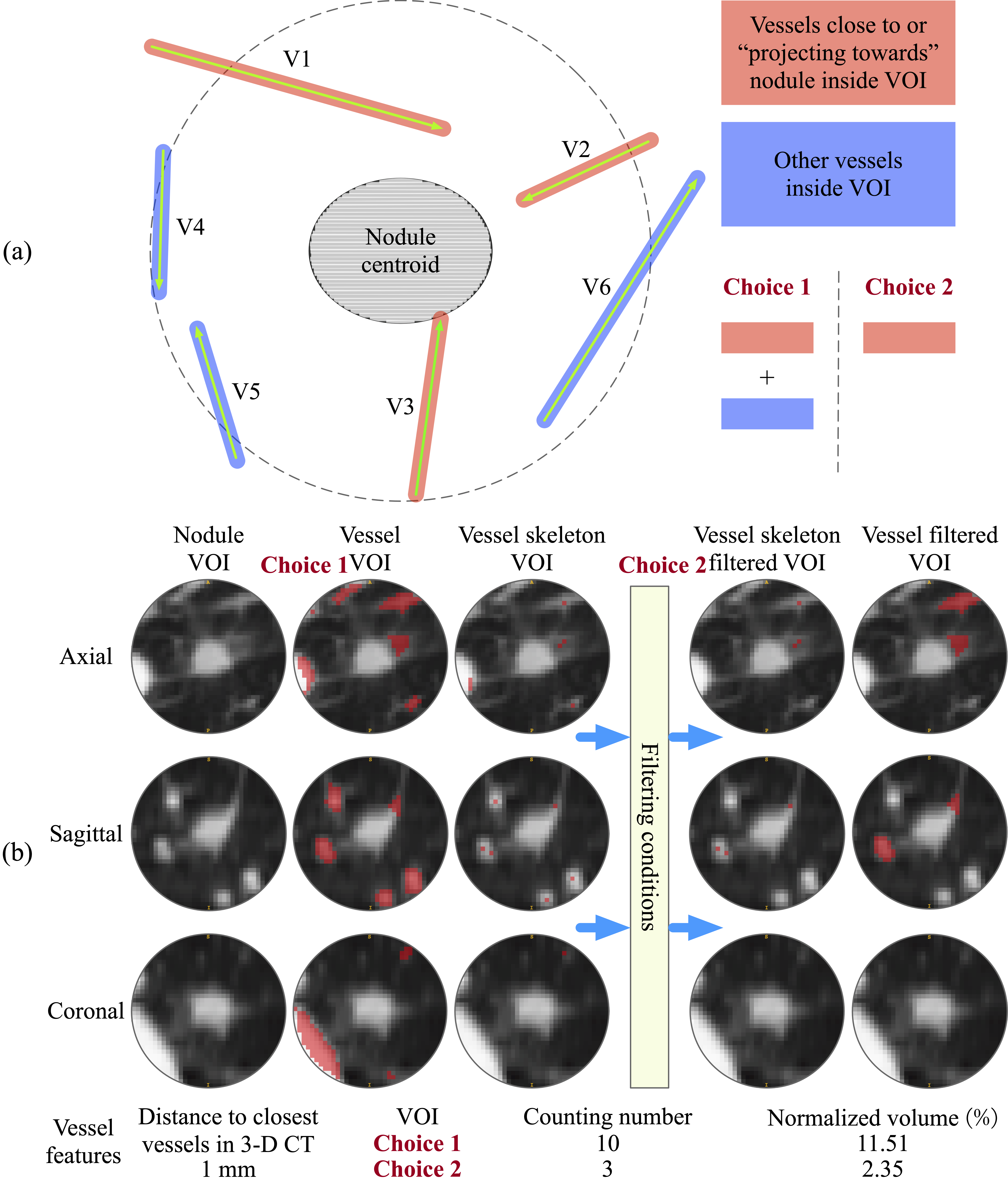}
\caption{Differences between choice 1 and 2 in identifying vessels inside VOI. (a) In choice 1, all branches were counted in quantification. In choice 2, only V1, V2 and V3 were quantified since they respectively matched the condition ii, iii and i. (b) Vessels and skeletons before and after filtering. Vessel features include the distance, counting number and normalized volume.}
\label{fig:voifilter}
\end{figure}

%vessel
%to vessels
%The visualized

To quantify the surrounding airways and vessels, we had two choices: 1) all structures inside VOI are evaluated; 2) only structures that are connected to or project towards a nodule are evaluated. For the second choice, we referred to Ref. \cite{wang2017vasculature} for the definition of filtering rules in distance and orientation:
\begin{itemize}
\item[\textbf{i.\ \ }] the centerline of airway or vessel is directly attached to a nodule;
\item[\textbf{ii.\ }] the distance between the centerline and the nodule surface is $\leq$ 3 mm;
\item[\textbf{iii.}] the distance between the centerline and the nodule surface is $\leq$ 5 mm and the centerline is projecting towards the nodule centroid.
\end{itemize}

%in a nodule VOI
%adjacent
%. Specifically, 
%conditions are defined as the following:

Here, ``projecting towards" means the angle, between centerline trajectory and the segment between a centerline end and the nodule centroid, is $\leq$ 15 degrees. All conditions above were summarized from clinical practice, but they might neglect some pertinent pulmonary structures. Therefore, both the two choices were adopted for quantification and comparison. Fig. \ref{fig:voifilter} illustrates the differences between choice 1 and choice 2 in identifying vessels.

%Although the filtering conditions were summarized from clinical knowledge\cite{wang2017vasculature}, the second choice 
%seemingly irrelevant
%Peripheral structures inside VOI might be pertinent as well.
%(e.g., axial slice), and the conditions above may not be accurate without 3-D visualization.  
%This is because radiologists usually analyze 3-D structures on 2-D plane 
%compared in our experiments
% in VOI

The quantification of airways and vessels also includes counting their number and volume inside VOI. Different from Ref. \cite{wang2017vasculature}, we divided the volume of structures by the volume of non-nodule region in VOI. Such normalized volume avoids the influence of nodule size. By nature, the VOI of a large nodule contains more airways and vessels. If the absolute volume is used, the volume increase of structures may be simply due to the enlarged VOI. Only through normalization can we properly validate the correlation between volume of surrounding structures and nodule malignancy.

%For airways and vessels, detailed quantification includes 
%respectively
%airways and vessels
%it is difficult to figure out whether
% by large nodule or related to nodule malignancy
% on statistical analysis

\subsection{Data analysis}

Data analysis was performed using Python with Scipy library \cite{virtanen2020scipy}.

%In analysis, a p-value $<$ 0.05 represents statistical significance.

\paragraph{Statistical analysis} The student's t-tests were conducted for comparisons of quantified structures between benign and malignant nodule groups. Correlation between features of pleurae, airways and vessels was tested with Pearson correlation coefficient (PCC), where two-tailed p-value was estimated. To interpret the relationship between pulmonary structures and nodules, features of surrounding structures were dichotomized with respect to nodule categories. Odds ratio (OR) and Chi-square tests ($\chi^2$) were performed on the resulting contingency tables.

%To intuitively interpret the relationship between pulmonary structures and nodule attribute, the distance to nodule surface, counting number and normalized volume are dichotomized. Then, OR and $\chi^2$ testing are performed on the corresponding contingency tables. 
% between dichotomized features of  and nodule categories in
%For each feature, the dichotomization threshold was obtained by considering both its mean value and clinical knowledge. 

\paragraph{Logistic regression} For each structure, a classifier was trained using quantified features. The model was evaluated on nodules from the testing set. ROC analysis was performed on the patient-level malignancy prediction results.

%Classification results of benignity-malignancy at patient-level were used for ROC analysis.
%as inputs
%logistic regression

\section{Results}

Both Table \ref{table:distribution1} and Table \ref{table:distribution2} showed that there existed significant differences between benign and malignant nodules in equivalent diameter and scoring disagreement among radiologists. The average diameters of benign and malignant nodules were respectively 6.97 mm and 14.52 mm. Malignant nodules tended to grow larger and lack expert consensus. Additionally, there were not significant differences between benign and malignant groups in diagnosed lung diseases such as emphysema and fibrosis. Pulmonary congestion was relatively frequently observed in patients with malignant nodules.

% The disagreement level among radiologists and the number of solid nodules over part-solid nodules are higher in the malignant group. 
% Significant differences in equivalent diameter and scorings are observed between benign and malignant groups (see ).
% the number of solid and part-solid nodules and .
%equivalent

\begin{table}[htbp]
\centering
\caption{Comparison of features of pleurae, airways and vessels surrounding the nodule}
\label{table:comp1}
%\begin{adjustbox}{width=\columnwidth}
\begin{threeparttable}
\begin{tabular}{llll}
\hline
Features & Benign & Malignant & All \\ \hline
Distance to nodule (mm)  &  &  &  \\
%\begin{tabular}[c]{@{}l@{}}Benign\\ scoring$<$3\end{tabular}  \begin{tabular}[c]{@{}l@{}}Malignant\\ scoring$>$3\end{tabular}
\quad pleural surface\tnote{$\dagger$}    &  6.56$\pm$6.93  & 5.19$\pm$6.09  &  6.11$\pm$6.69  \\
\quad airways\tnote{$\dagger$}   &  37.08$\pm$19.29  &  26.43$\pm$21.28  &  33.57$\pm$20.59  \\ 
\quad vessels\tnote{$\dagger$}   & 1.42$\pm$1.39  & 1.07$\pm$0.48  &  1.31$\pm$1.18 \\ 
\quad arteries\tnote{$\dagger$}   & 2.53$\pm$2.14  & 1.32$\pm$0.88 &  2.13$\pm$1.91 \\ 
\quad veins\tnote{$\dagger$}   & 2.25$\pm$2.42 & 1.32$\pm$0.94 & 1.94$\pm$2.10 \\ 
Counting number &  &  &  \\ 
\quad airways (choice 1)\tnote{$\dagger$} & 0.05$\pm$0.27 & 0.57$\pm$1.08  &  0.22$\pm$0.70 \\
\quad airways (choice 2)\tnote{$\dagger$} & 0.05$\pm$0.25 & 0.49$\pm$0.90  & 0.19$\pm$0.59 \\
\quad vessels (choice 1)\tnote{$\dagger$} & 6.63$\pm$9.79  &  34.94$\pm$39.09  &  15.94$\pm$27.28 \\
\quad vessels (choice 2)\tnote{$\dagger$} & 2.42$\pm$1.95  &  5.74$\pm$4.68  &  3.52$\pm$3.49 \\
\quad arteries (choice 1)\tnote{$\dagger$} & 5.02$\pm$10.17  &  31.95$\pm$38.25   &  13.88$\pm$26.66 \\
\quad arteries (choice 2)\tnote{$\dagger$} & 2.47$\pm$4.41  &  5.90$\pm$9.00  &  3.60$\pm$6.50 \\
\quad veins (choice 1)\tnote{$\dagger$} & 4.07$\pm$7.88  &  26.36$\pm$31.98  &  11.40$\pm$22.09 \\
\quad veins (choice 2)\tnote{$\dagger$} & 1.93$\pm$3.27  & 4.64$\pm$4.89  & 2.82$\pm$4.08 \\
Normalized volume (\%)   &  &  &  \\
\quad airways (choice 1)\tnote{$\dagger$}  &  0.09$\pm$0.72 &  0.29$\pm$0.88  & 0.16$\pm$0.78  \\
\quad airways (choice 2)\tnote{$\dagger$}  &  0.09$\pm$0.72 &  0.27$\pm$0.84  & 0.15$\pm$0.77  \\
\quad vessels (choice 1)\tnote{$\dagger$}  & 3.78$\pm$3.37 & 5.35$\pm$3.10 & 4.29$\pm$3.37  \\
\quad vessels (choice 2)\tnote{$\dagger$}  & 2.75$\pm$3.15 & 3.62$\pm$2.93 & 3.03$\pm$3.11  \\
\quad arteries (choice 1)\tnote{$\dagger$}  & 1.97$\pm$2.51 & 2.88$\pm$1.84 & 2.27$\pm$2.35 \\
\quad arteries (choice 2)  & 1.56$\pm$2.34 & 1.65$\pm$1.59  & 1.59$\pm$2.12 \\
\quad veins (choice 1)\tnote{$\dagger$}  & 1.77$\pm$2.01 & 2.39$\pm$1.72 & 1.98$\pm$1.94 \\
\quad veins (choice 2)  & 1.33$\pm$1.87 & 1.34$\pm$1.55 & 1.34$\pm$1.77 \\
\hline
\end{tabular}
\begin{tablenotes}
\item[$\dagger$]\footnotesize Statistical significance (p-value$<$0.05) was observed between different nodule groups.
\end{tablenotes}
\end{threeparttable}
%\end{adjustbox}
%in VOI
%in VOI
\end{table}

Table \ref{table:comp1} reflected that the average distance from nodule to pleural surface, airways, vessels, arteries and veins was significantly smaller in the malignant group. The counting number of all surrounding structures was different between benign and malignant groups. Similar results were observed in the normalized volume except for arteries and veins (choice 2). More structures were present around malignant nodules.

%Table \ref{table:comp1} provides quantified features of pulmonary structures surrounding nodules. 
%than in benign ones
%in the VOI of
% in terms of 

\begin{table}
\centering
\caption{Correlation among features of surrounding pulmonary structures and nodules}
\label{table:comp2}
\begin{adjustbox}{width=0.9\columnwidth}
\begin{threeparttable}
\begin{tabular}{llll}\hline
\multirow{2}{*}{Features} & \multicolumn{3}{c}{PCC (r)} \\
& Benign & Malignant & All \\ \hline
 % & \begin{tabular}[c]{@{}l@{}}Benign\\ scoring$<$3\end{tabular} & \begin{tabular}[c]{@{}l@{}}Malignant\\ scoring$>$3\end{tabular}
Pleurae  &  &  & \\
\quad Distance \& Eq. diameter & -0.08\tnote{$\dagger$}  &  -0.29\tnote{$\dagger$} &  -0.15\tnote{$\dagger$} \\
Airways  &  &  & \\
\quad Distance \& Eq. diameter & -0.14\tnote{$\dagger$}  &  -0.23\tnote{$\dagger$} &  -0.25\tnote{$\dagger$} \\
\quad Counting number \& Eq. diameter (choice 1) &  0.34\tnote{$\dagger$}  & 0.41\tnote{$\dagger$} & 0.50\tnote{$\dagger$} \\
\quad Counting number \& Eq. diameter (choice 2) &  0.14\tnote{$\dagger$} & 0.42\tnote{$\dagger$}   & 0.51\tnote{$\dagger$} \\
\quad Normalized volume \& Eq. diameter (choice 1) & 0.14\tnote{$\dagger$}  & 0.14\tnote{$\dagger$}  & 0.16\tnote{$\dagger$} \\
\quad Normalized volume \& Eq. diameter (choice 2) & 0.16\tnote{$\dagger$}  & 0.14\tnote{$\dagger$}  & 0.16\tnote{$\dagger$} \\
\quad Counting number \& normalized volume (choice 1) & 0.52\tnote{$\dagger$}  &  0.46\tnote{$\dagger$} &  0.40\tnote{$\dagger$} \\
\quad Counting number \& normalized volume (choice 2) &  0.53\tnote{$\dagger$}  &  0.43\tnote{$\dagger$}   & 0.39\tnote{$\dagger$}   \\
Vessels  &  &  &  \\
\quad Distance \& Eq. diameter & -0.10\tnote{$\dagger$}  &  -0.15\tnote{$\dagger$} &  -0.14\tnote{$\dagger$} \\
\quad Counting number \& Eq. diameter (choice 1) & 0.71\tnote{$\dagger$} & 0.63\tnote{$\dagger$} & 0.73\tnote{$\dagger$} \\
\quad Counting number \& Eq. diameter (choice 2) & 0.29\tnote{$\dagger$}  &  0.51\tnote{$\dagger$}  & 0.56\tnote{$\dagger$} \\
\quad Normalized volume \& Eq. diameter (choice 1) & 0.09\tnote{$\dagger$} &  -0.08   &  0.13\tnote{$\dagger$}\\
\quad Normalized volume \& Eq. diameter (choice 2) & 0.05 & -0.03 & 0.07\tnote{$\dagger$}\\
\quad Counting number \& normalized volume (choice 1)  &  0.15\tnote{$\dagger$}  &  0.04  & 0.15\tnote{$\dagger$} \\
\quad Counting number \& normalized volume (choice 2)  &  0.14\tnote{$\dagger$}   &  0.06   & 0.13\tnote{$\dagger$}  \\
Arteries   & &  & \\
\quad Distance \& Eq. diameter & -0.21\tnote{$\dagger$}  &  -0.31\tnote{$\dagger$} &  -0.28\tnote{$\dagger$} \\
\quad Counting number \& Eq. diameter (choice 1) & 0.71\tnote{$\dagger$} &  0.64\tnote{$\dagger$}  & 0.73\tnote{$\dagger$}\\
\quad Counting number \& Eq. diameter (choice 2) & 0.30\tnote{$\dagger$} & 0.23\tnote{$\dagger$} & 0.34\tnote{$\dagger$}\\
\quad Normalized volume \& Eq. diameter (choice 1) & 0.08\tnote{$\dagger$} & -0.06 & 0.12\tnote{$\dagger$}\\
\quad Normalized volume \& Eq. diameter (choice 2) & 0.01 & -0.17\tnote{$\dagger$}   & -0.02\\
\quad Counting number \& normalized volume (choice 1) & 0.13\tnote{$\dagger$}  &  0.06  &  0.14\tnote{$\dagger$} \\
\quad Counting number \& normalized volume (choice 2) &  0.09\tnote{$\dagger$}  &  0.09\tnote{$\dagger$}  &  0.08\tnote{$\dagger$} \\
Veins   & & & \\
\quad Distance \& Eq. diameter & -0.11\tnote{$\dagger$}  &  -0.27\tnote{$\dagger$} &  -0.19\tnote{$\dagger$} \\
\quad Counting number \& Eq. diameter (choice 1) & 0.74\tnote{$\dagger$} & 0.64\tnote{$\dagger$}  & 0.73\tnote{$\dagger$} \\
\quad Counting number \& Eq. diameter (choice 2) & 0.53\tnote{$\dagger$} & 0.35\tnote{$\dagger$}  & 0.48\tnote{$\dagger$} \\
\quad Normalized volume \& Eq. diameter (choice 1) & 0.04 & -0.09\tnote{$\dagger$} & 0.06\tnote{$\dagger$} \\
\quad Normalized volume \& Eq. diameter (choice 2) & -0.02 &  -0.21\tnote{$\dagger$}  & -0.09\tnote{$\dagger$} \\
\quad Counting number \& normalized volume (choice 1) & 0.14\tnote{$\dagger$}  & 0.04  & 0.12\tnote{$\dagger$}  \\
\quad Counting number \& normalized volume (choice 2) &  0.12\tnote{$\dagger$}  &  0.03  &  0.09\tnote{$\dagger$} \\
\hline
\end{tabular}
\begin{tablenotes}
\item[$\dagger$]\footnotesize Low probability (p-value$<$0.05) was observed if the two feature variables were uncorrelated.
\end{tablenotes}
\end{threeparttable}
\end{adjustbox}
\end{table}

The distance from nodule to pleurae, airways and vessels was negatively associated with nodule diameter, but the correlation was weak (see Table \ref{table:comp2}). For vessels, arteries and veins around benign nodules, the correlation between counting number and normalized volume was weak. Such correlation was weaker for malignant nodules. The counting number of all surrounding structures was strongly correlated with diameter.

% Correlation analysis  demonstrates that 
% among features of surrounding structures and nodule diameter 
% with PCC around 0.1--0.15
% with PCC ranging from 0.03 to 0.09.
%inside VOI
%In comparison with normalized volume, 

\begin{table}
\centering
\caption{Relationship between dichotomized distances to structures and nodules}
\label{table:comp3}
\begin{adjustbox}{width=\columnwidth}
%\begin{threeparttable}
\begin{tabular}{llllllll}
\hline
\multicolumn{2}{l}{Distance to} & \multicolumn{2}{l}{Malignancy} & \multicolumn{2}{l}{Eq. diameter} & \multicolumn{2}{l}{Texture} \\
%\multirow{2}{*}{}\tnote{$\dagger$}
\multicolumn{2}{l}{nodule (mm)} & Benign & Malignant & $<$10 mm & $\geq$10 mm & Solid & Part-solid \\ \hline
\multirow{2}{*}{Pleurae} & $\leq$1 &  407  &  253 &  400 &  260 & 475 & 185 \\
 & $>$1 & 637  & 259 & 708  &  188 & 615 & 281 \\
\multirow{2}{*}{Airways} & $\leq$1 & 12 & 50 & 14 & 48  & 47  & 15  \\
 & $>$1 & 1032 & 462  & 1094  & 400  & 1043 & 451 \\
\multirow{2}{*}{Vessels} & $\leq$1 & 896 & 498 & 950  & 444  & 966 & 428 \\
 & $>$1 & 148 & 14  & 158 & 4 & 124 & 38 \\
\multirow{2}{*}{Arteries} & $\leq$1 & 540 & 433 & 572  & 401 & 633 & 340 \\
 & $>$1 & 504 &  79 & 536 & 47 & 457  & 126 \\
\multirow{2}{*}{Veins} & $\leq$1 & 681 & 432  &  711 & 402 & 780 & 333 \\
 & $>$1 & 363 & 80  &  397 & 46 & 310 & 133  \\
\hline
\end{tabular}
%\begin{tablenotes}
%\item[$\dagger$]\footnotesize Here, distance$\leq$1 mm indicates that a nodule is attached to certain structures.
%\end{tablenotes}
%\end{threeparttable}
\end{adjustbox}
\end{table}

\begin{table}
\centering
\caption{Relationship between dichotomized counting number of structures and nodules}
\label{table:comp4}
\begin{adjustbox}{width=\columnwidth}
\begin{tabular}{llllllll}
\hline
\multicolumn{2}{l}{Counting} & \multicolumn{2}{l}{Malignancy} & \multicolumn{2}{l}{Eq. diameter} & \multicolumn{2}{l}{Texture} \\
\multicolumn{2}{l}{number} & Benign & Malignant & $<$10 mm & $\geq$10 mm & Solid & Part-solid \\ \hline
\multirow{2}{*}{\begin{tabular}[c]{@{}l@{}}Airways\\ (choice 1)\end{tabular}} & $\leq$1 & 1038 & 436 & 1104 & 370  & 1035  & 439  \\
 & $>$1 & 6 & 76  & 4  & 78  & 55 & 27 \\
\multirow{2}{*}{\begin{tabular}[c]{@{}l@{}}Airways\\ (choice 2)\end{tabular}} & $\leq$1 & 1038   & 452  & 1105  & 385    &  1048  &  442  \\
 & $>$1 & 6   &  60    &  3  & 63   & 42   &  24  \\
\multirow{2}{*}{\begin{tabular}[c]{@{}l@{}}Vessels\\ (choice 1)\end{tabular}} & $\leq$10 & 907 & 108 & 966  & 49  & 733 & 282 \\
 & $>$10 & 137 & 404  & 142 & 399 & 357 & 184 \\
\multirow{2}{*}{\begin{tabular}[c]{@{}l@{}}Vessels\\ (choice 2)\end{tabular}} & $\leq$3 & 850  & 196  & 910   &  136  & 748  & 298  \\
 & $>$3 & 194   &  316   & 198  & 312   & 342  & 168  \\
\multirow{2}{*}{\begin{tabular}[c]{@{}l@{}}Arteries\\ (choice 1)\end{tabular}} & $\leq$10 & 964 & 144 & 1024  & 84 & 788 & 320 \\
 & $>$10 & 80 &  368 & 84 & 364 & 302  & 146 \\
\multirow{2}{*}{\begin{tabular}[c]{@{}l@{}}Arteries\\ (choice 2)\end{tabular}} & $\leq$3 &  839  & 232  &  879  & 192  & 755  & 316  \\
 & $>$3 & 205   & 280   &  229  & 256  &  335  &  150 \\
\multirow{2}{*}{\begin{tabular}[c]{@{}l@{}}Veins\\ (choice 1)\end{tabular}} & $\leq$10 & 991 & 174  &  1060 & 105 & 827 & 338 \\
 & $>$10 & 53 & 338  &  48 & 343 & 263 & 128  \\
\multirow{2}{*}{\begin{tabular}[c]{@{}l@{}}Veins\\ (choice 2)\end{tabular}} & $\leq$3 &  916  &  280   &  979  & 217  & 849  &  347 \\
 & $>$3 &  128  &  232  & 129   & 231  &  241  &  119  \\
\hline
\end{tabular}
\end{adjustbox}
\end{table}

\begin{table}
\centering
\caption{Relationship between dichotomized normalized volume of structures and nodules}
\label{table:comp5}
\begin{adjustbox}{width=\columnwidth}
\begin{tabular}{llllllll}
\hline
\multicolumn{2}{l}{Normalized} & \multicolumn{2}{l}{Malignancy} & \multicolumn{2}{l}{Eq. diameter} & \multicolumn{2}{l}{Texture} \\
\multicolumn{2}{l}{volume (\%)} & Benign & Malignant & $<$10 mm & $\geq$10 mm & Solid & Part-solid \\ \hline
\multirow{2}{*}{\begin{tabular}[c]{@{}l@{}}Airways\\ (choice 1)\end{tabular}} & $\leq$0.1 & 1010 & 391 & 1074 & 327  & 978  & 423  \\
 & $>$0.1 & 34 & 121  & 34  & 121  & 112 & 43 \\
 \multirow{2}{*}{\begin{tabular}[c]{@{}l@{}}Airways\\ (choice 2)\end{tabular}} & $\leq$0.1 & 1010  &  391 &  1074  &  327   &  978  & 423   \\
 & $>$0.1 & 34   &   121  & 34   &  121  &  112  & 43   \\
\multirow{2}{*}{\begin{tabular}[c]{@{}l@{}}Vessels\\ (choice 1)\end{tabular}} & $\leq$2 & 365 & 44 & 379  & 30  & 296 & 113 \\
 & $>$2 & 679 & 468  & 729 & 418 & 794 & 353 \\
\multirow{2}{*}{\begin{tabular}[c]{@{}l@{}}Vessels\\ (choice 2)\end{tabular}} & $\leq$2 & 563  &  173  & 585   & 151   &  514  & 222  \\
 & $>$2 &  481  & 339    &  523  &  297   &  576  & 244  \\
\multirow{2}{*}{\begin{tabular}[c]{@{}l@{}}Arteries\\ (choice 1)\end{tabular}} & $\leq$2 & 704 & 181 & 721  & 164 & 639 & 246 \\
 & $>$2 & 340 &  331 & 387 & 284 & 451  & 220 \\
\multirow{2}{*}{\begin{tabular}[c]{@{}l@{}}Arteries\\ (choice 2)\end{tabular}} & $\leq$2 &  796  & 357  &  821  & 332  & 814  & 339  \\
 & $>$2 & 248   & 155   & 287   &  116 & 276   & 127  \\
\multirow{2}{*}{\begin{tabular}[c]{@{}l@{}}Veins\\ (choice 1)\end{tabular}} & $\leq$2 & 689 & 254  &  715 & 228 & 654 & 289 \\
 & $>$2 & 355 & 258  &  393 & 220 & 436 & 177  \\
\multirow{2}{*}{\begin{tabular}[c]{@{}l@{}}Veins\\ (choice 2)\end{tabular}} & $\leq$2 &  810  &  408   &  846  &  372 & 854   & 364  \\
 & $>$2  &  234  & 104   & 262   & 76  &  236  &  102  \\
\hline
\end{tabular}
\end{adjustbox}
\end{table}

%To intuitively interpret the relationship between pulmonary structures and nodule attribute, the distance to nodule surface, counting number and normalized volume are dichotomized. Then, OR and $\chi^2$ testing are performed on the corresponding contingency tables. 

%Tables \ref{table:comp3}, \ref{table:comp4} and \ref{table:comp5}. Results of OR and  are presented below, where $\dagger$ denotes p-value$<$0.05.
%surrounding a nodule

Table \ref{table:corr_distance} provided the correlation testing results of Table \ref{table:comp3}. The attachment to pleurae, airways, vessels and veins was hardly relevant to nodule texture. The attachment to arteries was more often observed in part-solid nodules. Nodules contacting pleural surface and tubular structures were less likely to be benign and small. 

%Nodules adjacent to tubules are relatively frequent in malignant and large groups.
%Nodules respectively contacting  are more often seen in malignant and large groups.

\begin{table}
\centering
\caption{Correlation testing between dichotomized distance and nodule attribute}
\label{table:corr_distance}
%\begin{adjustbox}{width=\columnwidth}
\begin{threeparttable}
\begin{tabular}{llll}
\hline
%\multicolumn{2}{l}{\end{tabular}}
Distance to nodule (mm)  & Nodule attribute  &  OR  &  $\chi^2$  \\ \hline
\multirow{3}{*}{Pleurae} & Malignancy & 0.65$\dagger$  &  15.30$\dagger$  \\
 & Eq. diameter &  0.41$\dagger$  &  62.84$\dagger$  \\
 & Texture &  1.17  &  2.01  \\
\multirow{3}{*}{Airways} & Malignancy &  0.11$\dagger$  &   66.66$\dagger$   \\
 & Eq. diameter &   0.11$\dagger$  &   74.48$\dagger$   \\
 & Texture &   1.35  &  1.02   \\
\multirow{3}{*}{Vessels} & Malignancy &   0.17$\dagger$  &  48.22$\dagger$    \\
 & Eq. diameter &   0.05$\dagger$  &   61.11$\dagger$  \\
 & Texture &  0.69   &   3.63  \\
\multirow{3}{*}{Arteries} & Malignancy &   0.20$\dagger$   &   158.19$\dagger$  \\
 & Eq. diameter &  0.13$\dagger$  &  195.42$\dagger$   \\
 & Texture  &   0.51$\dagger$  &   30.88$\dagger$     \\
\multirow{3}{*}{Veins} & Malignancy &   0.35$\dagger$  &  61.83$\dagger$  \\
 &  Eq. diameter &  0.20$\dagger$  &   102.36$\dagger$  \\ 
 & Texture  &  1.00  &  0.00  \\  \hline
\end{tabular}
\begin{tablenotes}
\item[$\dagger$] \footnotesize Statistical significance (p-value$<$0.05) was observed between different nodule groups.
\end{tablenotes}
%\end{adjustbox}
\end{threeparttable}
\end{table}

Table \ref{table:corr_number} provided the correlation testing results of Table \ref{table:comp4}. The counting number of all structures except vessels (choice 2) was irrelevant to nodule texture. The more tubule branches exist inside VOI, the more likely the nodule is malignant.

%Detailed correlation testing results are presented in Table \ref{table:corr_number}.
%for airways, vessels (choice 2), arteries and veins. The number of vessels (choice 1) is slightly relevant to texture, but that in choice 2 is irrelevant. 
%The quantity of tubules is positively related to nodule malignancy and diameter. 

\begin{table}
\centering
\caption{Correlation testing between dichotomized counting number and nodule attribute}
\label{table:corr_number}
%\begin{adjustbox}{width=\columnwidth}
\begin{threeparttable}
\begin{tabular}{llllll}
\hline
Counting & Nodule & \multicolumn{2}{l}{Choice 1} & \multicolumn{2}{l}{Choice 2} \\
number & attribute & OR & $\chi^2$ & OR & $\chi^2$ \\ \hline
\multirow{3}{*}{Airways} & Malignancy &   30.16$\dagger$   &  140.11$\dagger$    &   22.96$\dagger$   &   105.04$\dagger$    \\
 & Eq. diameter &  58.18$\dagger$    &   185.76$\dagger$  &  60.27$\dagger$   &  149.39$\dagger$   \\
 & Texture &  1.16   &   0.37  &  1.35  &  1.35 \\
\multirow{3}{*}{Vessels} & Malignancy &   24.77$\dagger$  &   655.47$\dagger$   &   7.06$\dagger$  &  290.11$\dagger$   \\
 & Eq. diameter &   55.39$\dagger$  &  817.72$\dagger$  &  10.54$\dagger$  &  388.09$\dagger$ \\
 & Texture &  1.34$\dagger$  &  6.52$\dagger$  &  1.23  &  3.24  \\
\multirow{3}{*}{Arteries} & Malignancy &  30.79$\dagger$  &  690.87$\dagger$  &  4.94$\dagger$  &  196.73$\dagger$   \\
 & Eq. diameter &  52.83$\dagger$  &   844.45$\dagger$  &  5.12$\dagger$  &  197.83$\dagger$  \\
 & Texture  &  1.19  &  2.09  &  1.07  &  0.32    \\
\multirow{3}{*}{Veins} & Malignancy &  36.32$\dagger$  &  678.06$\dagger$  &   5.93$\dagger$  &  211.03$\dagger$   \\
 &  Eq. diameter &  72.14$\dagger$  &  884.64$\dagger$  &   8.08$\dagger$  &  285.87$\dagger$  \\ 
 & Texture  &  1.19$\dagger$  &  1.93$\dagger$  &  1.21  &  2.16   \\  \hline
\end{tabular}
\begin{tablenotes}
\item[$\dagger$] \footnotesize Statistical significance (p-value$<$0.05) was observed between different nodule groups.
\end{tablenotes}
\end{threeparttable}
%\begin{tabular}{llll}
%\hline
%\multicolumn{2}{l}{Counting number} & Choice 1 & Choice 2 \\ \hline
%\multirow{2}{*}{\begin{tabular}[c]{@{}l@{}} \\ airways\\  \end{tabular}} & Malignancy & (OR=30.16$\dagger$, $\chi^2$=140.11$\dagger$) & (OR=22.96$\dagger$, $\chi^2$=105.04$\dagger$) \\
% & \begin{tabular}[c]{@{}l@{}}Equivalent\\ diameter\end{tabular} & (OR=58.18$\dagger$, $\chi^2$=185.76$\dagger$) & (OR=60.27$\dagger$, $\chi^2$=149.39$\dagger$) \\
%\multirow{2}{*}{\begin{tabular}[c]{@{}l@{}} \\ vessels \\ \end{tabular}} & Malignancy & (OR=24.77$\dagger$, $\chi^2$=655.47$\dagger$) & (OR=7.06$\dagger$, $\chi^2$=290.11$\dagger$) \\
% & \begin{tabular}[c]{@{}l@{}}Equivalent\\ diameter\end{tabular} & (OR=55.39$\dagger$, $\chi^2$=817.72$\dagger$) & (OR=10.54$\dagger$, $\chi^2$=388.09$\dagger$) \\
%\multirow{2}{*}{\begin{tabular}[c]{@{}l@{}} \\ arteries\\ \end{tabular}} & Malignancy & (OR=30.79$\dagger$, $\chi^2$=690.87$\dagger$) & (OR=4.94$\dagger$, $\chi^2$=196.73$\dagger$) \\
% & \begin{tabular}[c]{@{}l@{}}Equivalent\\ diameter\end{tabular} & (OR=52.83$\dagger$, $\chi^2$=844.45$\dagger$) & (OR=5.12$\dagger$, $\chi^2$=197.83$\dagger$)  \\
%\multirow{2}{*}{\begin{tabular}[c]{@{}l@{}} \\ veins \\ \end{tabular}} & Malignancy & (OR=36.32$\dagger$, $\chi^2$=678.06$\dagger$) & (OR=5.93$\dagger$, $\chi^2$=211.03$\dagger$) \\
% & \begin{tabular}[c]{@{}l@{}}Equivalent\\ diameter\end{tabular} & (OR=72.14$\dagger$, $\chi^2$=884.64$\dagger$) & (OR=8.08$\dagger$, $\chi^2$=285.87$\dagger$) \\ \hline
%\end{tabular}
%\end{adjustbox}
\end{table}

Table \ref{table:corr_volume} provided the correlation testing results of Table \ref{table:comp5}. The normalized volume of all structures except artery (choice 1) was irrelevant to nodule texture. Except for arteries and veins (choice 2), the volume of all surrounding structures was positively related to nodule malignancy and diameter. The larger percentage of VOI these structures occupy, the more likely the nodule is malignant and large.

%Detailed correlation testing results are given in .
% for airways, vessels and veins. The artery volume of choice 1 is slightly relevant to texture, but that of choice 2 is irrelevant. 

\begin{table}
\centering
\caption{Correlation testing between dichotomized normalized volume and nodule attribute}
\label{table:corr_volume}
%\begin{adjustbox}{width=\columnwidth}
\begin{threeparttable}
\begin{tabular}{llllll}
\hline
Normalized &  Nodule  & \multicolumn{2}{l}{Choice 1} & \multicolumn{2}{l}{Choice 2} \\
volume (\%) & attribute  & OR & $\chi^2$ & OR & $\chi^2$ \\ \hline
\multirow{3}{*}{Airways} & Malignancy &  9.19$\dagger$  &  159.02$\dagger$  &  9.19$\dagger$  &  159.02$\dagger$   \\
 & Eq. diameter & 11.69$\dagger$  &  203.85$\dagger$  &  11.69$\dagger$  &  203.85$\dagger$   \\
 & Texture   &   0.89  &   0.40   &  0.89  & 0.40  \\
\multirow{3}{*}{Vessels} & Malignancy &  5.72$\dagger$  &  123.27$\dagger$  &   2.29$\dagger$  &  55.89$\dagger$  \\
 & Eq. diameter &  7.24$\dagger$   &  124.60$\dagger$    &   2.20$\dagger$   &  46.65$\dagger$   \\
 & Texture  &   1.16  &  1.42   &   0.98  &  0.03   \\
\multirow{3}{*}{Arteries} & Malignancy &  3.79$\dagger$  &  144.15$\dagger$  &  1.39$\dagger$  &  7.61$\dagger$  \\
 & Eq. diameter &  7.24$\dagger$  &  124.60$\dagger$  &  1.00  &  0.00  \\
 & Texture  &  1.27$\dagger$  &   4.53$\dagger$ &    1.10  &  0.63  \\
\multirow{3}{*}{Veins} & Malignancy &   1.97$\dagger$  &  38.64$\dagger$   &  0.88  &  0.89  \\
 & Eq. diameter &  1.76$\dagger$  &  24.85$\dagger$  &  0.66$\dagger$  &  8.38$\dagger$   \\ 
& Texture   &   0.92  &  0.56    &   1.01   &  0.01   \\ \hline
\end{tabular}
\begin{tablenotes}
\item[$\dagger$] \footnotesize Statistical significance (p-value$<$0.05) was observed between different nodule groups.
\end{tablenotes}
\end{threeparttable}
%\begin{tabular}{llll}
%\hline
%\multicolumn{2}{l}{Normalized volume} & Choice 1 & Choice 2 \\ \hline
%\multirow{2}{*}{\begin{tabular}[c]{@{}l@{}}\\  airways\\ \end{tabular}} & Malignancy & (OR=9.19$\dagger$, $\chi^2$=159.02$\dagger$) & (OR=9.19$\dagger$, $\chi^2$=159.02$\dagger$) \\
% & \begin{tabular}[c]{@{}l@{}}Equivalent\\ diameter\end{tabular} & (OR=11.69$\dagger$, $\chi^2$=203.85$\dagger$) & (OR=11.69$\dagger$, $\chi^2$=203.85$\dagger$) \\
%\multirow{2}{*}{\begin{tabular}[c]{@{}l@{}} \\ vessels\\ \end{tabular}} & Malignancy & (OR=5.72$\dagger$, $\chi^2$=123.27$\dagger$) &  (OR=2.29$\dagger$, $\chi^2$=55.89$\dagger$) \\
% & \begin{tabular}[c]{@{}l@{}}Equivalent\\ diameter\end{tabular} & (OR=7.24$\dagger$, $\chi^2$=124.60$\dagger$) & (OR=2.20$\dagger$, $\chi^2$=46.65$\dagger$) \\
%\multirow{2}{*}{\begin{tabular}[c]{@{}l@{}} \\ arteries\\ \end{tabular}} & Malignancy & (OR=3.79$\dagger$, $\chi^2$=144.15$\dagger$) & (OR=1.39$\dagger$, $\chi^2$=7.61$\dagger$) \\
% & \begin{tabular}[c]{@{}l@{}}Equivalent\\ diameter\end{tabular} & (OR=7.24$\dagger$, $\chi^2$=124.60$\dagger$)  & (OR=1.00, $\chi^2$=0.00) \\
%\multirow{2}{*}{\begin{tabular}[c]{@{}l@{}} \\ veins\\ \end{tabular}} & Malignancy & (OR=1.97$\dagger$, $\chi^2$=38.64$\dagger$) & (OR=0.88, $\chi^2$=0.89) \\
% & \begin{tabular}[c]{@{}l@{}}Equivalent\\ diameter\end{tabular} & (OR=1.76$\dagger$, $\chi^2$=24.85$\dagger$) & (OR=0.66$\dagger$, $\chi^2$=8.38$\dagger$)  \\ \hline
%\end{tabular}
%\end{adjustbox}
\end{table}

\begin{figure}[htbp]
\centering
\includegraphics[width=\columnwidth]{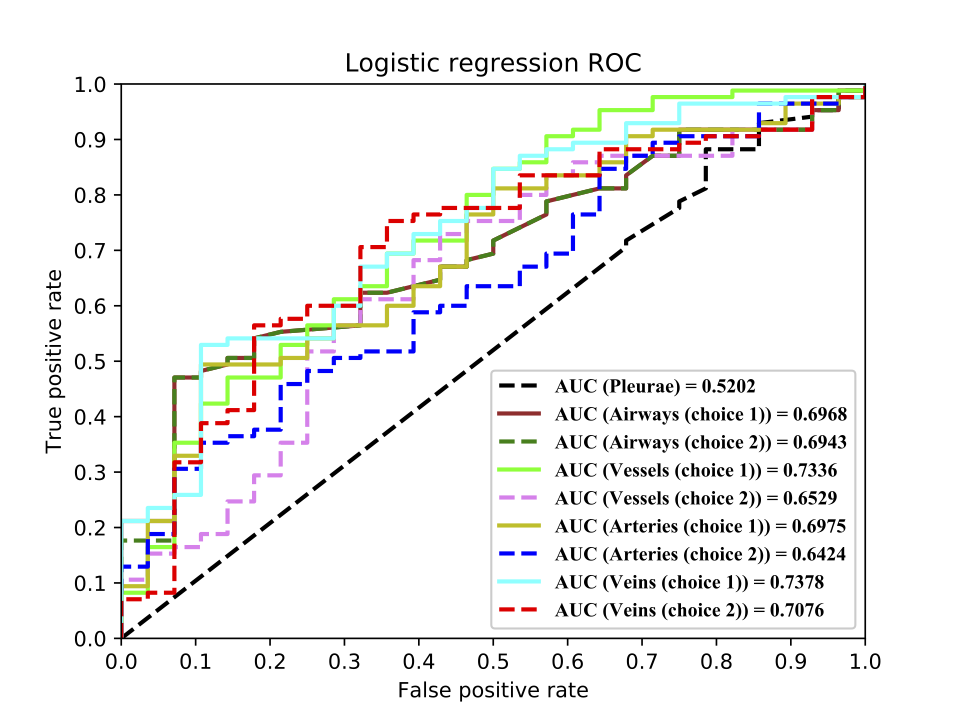}
\caption{ROC analysis of logistic regression. Features of surrounding pulmonary structures were used to discriminate benign and malignant nodules and predict patient-level malignancy.}
\label{fig:roc}
\end{figure}

Performance and ROC analysis of logistic regression were reported in Table \ref{table:comp6} and Fig. \ref{fig:roc}. The pleural feature only involved the distance to nodule. For other structures, the distance, counting number and normalized volume were combined as features. The artery features (choice 2) achieved the highest accuracy, recall and $F_1$-score while the vein features (choice 1) gained the highest areas-under-curve (AUC). The highest precision was obtained by airway features of both choices 1 and 2.

%Fig. \ref{fig:roc} plotted ROC curves for each structure.
%together
%in benign-malignant classification
%to pleural surface

\begin{table}
\centering
\caption{Performance of logistic regression for patient-level malignancy prediction}
\label{table:comp6}
%\begin{adjustbox}{width=\columnwidth}
\begin{tabular}{llllll}
\hline
Structure features & Accuracy & Precision & Recall & $F_1$-score &  AUC \\ \hline
Pleurae & 0.7080 &  0.7708  &  0.8706  & 0.8177  & 0.5202 \\
Airways (choice 1) & 0.5841 & \textbf{0.9524} & 0.4706  &  0.6299  &   0.6968 \\ 
Airways (choice 2) & 0.5841 & \textbf{0.9524}  & 0.4706  &  0.6299  &  0.6943 \\
Vessels (choice 1) & 0.6814 & 0.8451  &  0.7059  &  0.7692  &  0.7336 \\
Vessels (choice 2) & 0.6903 &  0.8125  &  0.7647  &  0.7879  &   0.6529 \\
Arteries (choice 1) & 0.6991 & 0.8228  & 0.7647  &  0.7927  &  0.6975 \\
Arteries (choice 2) & \textbf{0.7345}   &   0.7895  &  \textbf{0.8824}  &  \textbf{0.8333}  & 0.6424 \\
Veins (choice 1) & 0.6991 & 0.8400  & 0.7412  & 0.7875  &  \textbf{0.7378} \\
Veins (choice 2) & 0.7257 & 0.8553  &  0.7647  &  0.8075  &  0.7076 \\
\hline
\end{tabular}
%\end{adjustbox}
\end{table}

\section{Discussion}

Our study suggests that the distance, counting number and normalized volume of airways, vessels, arteries and veins can be viewed as important biomarkers for discriminating benign nodules from malignant ones. To the best of our knowledge, this is the first quantitative investigation using a large-scale publicly available chest CT dataset to discover the relation between pulmonary structures and nodule malignancy. It may provide new potential diagnosis basis for indeterminate nodules, and thereafter improve diagnostic accuracy and reduce unnecessary intervention. Our previously proposed segmentation methods make it possible to quantify structures from massive CT samples efficiently and accurately, which is a prerequisite to draw general conclusions.

%For diagnosis-definite cases in LIDC-IDRI, .
%Owing to our previously proposed segmentation methods, the efficiency and accuracy of labelling structures from CT were improved. We believe the quantification of structures on a massive dataset.
%Our goal is to . It may further  patient-level  and reduce follow-up procedures initiated during CT lung screening.
%the surrounding pleurae, airways and vessels are all informative in predicting nodule malignancy. 
%Specifically
%deep learning
%the labor of manually labeling structures from CT is avoided. 
%evaluate their discrimination ability  . 

\paragraph{Pleurae} The average distance from pleurae to malignant nodules is smaller than that to benign nodules. Meanwhile, the correlation between distance and nodule size is weak, revealing that: 1) the distance is not necessarily negatively related to nodule size because the position distribution of nodules is diverse; 2) the malignancy is indeed associated with the distance. After dichotomization, the malignant inclination of nodules that directly contact pleurae is clearer. The competitive accuracy versus low AUC suggests that the pleural feature is not robust enough on patient-level malignancy prediction and the binarization threshold needs to be tuned carefully.

%Logistic regression accuracy confirmed the effectiveness of such distance in patient-level malignancy prediction. Its low AUC suggests that the  otherwise the performance might drop significantly.
%From the distance to pleural surface, we find that 
%(Equivalent diameter)
%Malignant nodules are closer to pleurae.
%from nodule to pleurae
%of distance into direct contact or not,
%tendency

\paragraph{Airways} Most nodules in LIDC-IDRI dataset are not adjacent to airways. Both the counting number and normalized volume of airways are close to 0. Consequently, correlation between the two features is relatively high. Nearly 90\% of the nodules that have more than one airway branch around are malignant. The classifier may simply learn to predict a nodule as malignant provided that airways exist nearby. As a result, the airway feature achieved the highest precision. But for the remaining nodules without airways, the classifier retrieved few malignant ones with a rather low recall.

%The distance from a nodule to airways is over 33 mm on average, and only 62 out of 1556 nodules are attached to airways (wall). 
%The reason why logistic regression model using airway features achieved the highest precision can be explained as follows.
%Nodules that contact airway walls with diameter over 10 mm are highly likely to be malignant.
%when they . 
%numerous
%accompanying
%precision would be high. 

\paragraph{Vessels} Around 97\% of the malignant nodules are juxta-vascular and their average distance to vessels is 1.07 mm. Malignant nodules have more vessel branches than benign ones. The correlation between volume and nodule size is very weak, implying that the increase of vessel volume is not merely an outcome of the expanded VOI. The proportion of malignant nodules contacting vessels or having more than 10 branches in the vicinity is much higher than the opposite. Vessel features exhibited modest power to recognize malignant nodules. 

%much
% Such finding is consistent with Ref. , corroborating that 
%When it comes to vessel features, 
%, which is in line with the fact that 
%Correspondingly, the normalized volume of vessels is larger in the malignant group. 
%with respect to all nodules is 
%In logistic regression, 

\paragraph{Arteries and Veins} Similar to the findings of vessel features, the distance to nodule, the counting number and normalized volume of arteries and veins are respectively smaller and larger in the malignant group. But the correlation between counting number and nodule size is strong. Nodules surrounded by more arteries and veins tend to be malignant and large. Moreover, either artery or vein features performed well in patient-level malignancy discrimination.

%the average distance from a nodule to its surrounding arteries or veins is smaller in the malignant group. Besides, the average counting number and normalized volume of arteries or veins are larger for malignant nodules. 
%If vessels are further separated into arteries and veins, more comparison can be performed. 
%This finding is quite interesting because previous studies  concluded that in comparison with benign nodules, the counting number of veins, rather than that of arteries, was significantly greater in malignant nodules. Our findings are in accord with Gao et al.  that no significant difference was found between the presence of arteries and veins. 
% than in the benign group
%But the correlation among other features is weak. 
%for both arteries and veins. In dichotomized data,
%the logistic regression using 

\paragraph{Verification of previous findings} We compared findings in previous studies with ours. In this study, there were 103 benign, 8 malignant, and 99 uncertain non-calcified solid nodules attached to the costal pleura that have smooth margins with diameters less than 10 mm. Such distribution is a bit different from Ref. \cite{zhu2020management} where all these nodules were benign. Nevertheless, the proportion of benign nodules is still much larger. Consistent with Ref. \cite{wang2017vasculature}, vessels surrounding a nodule did play a role in benign-malignant distinction. Another interesting finding is that there exists no significant difference between the presence of arteries and veins around malignant nodules, which is in accord with Ref. \cite{gao2013multi} but against Ref. \cite{mori1990small}. Since only 26 patients were investigated by Mori et al. \cite{mori1990small}, their conclusion might be biased due to limited sample size.

%, oval, semi-circular or triangular shapes
%they found
%than that of malignant ones
%experimental results corroborated that we believe 
% (counting number)

\paragraph{Choice 1 or 2} The difference between choice 1 and 2 lies in the filtering of target structures, but such difference did not lead to contradictory findings. Features of vessels, arteries and veins in choice 2 outperformed those in choice 1 in accuracy, recall and $F_1$-score. The vessels (choice 2) are more closely related to nodule malignancy, corroborating the value of filtering weakly correlated structures.

%Difference between choices 1 and 2
%In fact, consistent conclusions can be drawn from results of statistical analyses and logistic regression.
%classification metrics such as
%, which have attachment to or project towards nodules, 
% after filtering
%Our experiment results suggest that it is more appropriate to analyze filtered target structures in clinical practice.
% in clinical practice

\paragraph{Limitations} First, some low-resolution CT scans in LIDC-IDRI restrict the extent to which the details of structures are segmented. Errors accumulate along the pipeline of segmentation and quantification. More high-quality scans should be collected. Second, the number of pathologically-proved nodules is insufficient. Subjective malignancy scorings cannot substitute for pathological ground-truth. On the other hand, percutaneous biopsy is usually carried out on one highly-suspicious malignant nodule (solid, diameter over 15 mm) for patients with multiple nodules \cite{anzidei2017imaging, kothary2009computed}. 

Considering that benign-definite and malignant-definite patients under investigation respectively had 2.33 and 2.86 nodules on average, it is laborious or even impractical to obtain labels of all nodules in each patient. Therefore, it takes time to collect more nodule samples with ground-truth. In the future, the effectiveness of structure features as lung cancer biomarkers could be further evaluated. Third, more elaborate features could be considered in quantification, which may improve the discrimination ability of pulmonary structures on nodule malignancy.

%needs to be further evaluated in the future with more collected samples.
%In the future, more nodule samples with ground-truth will be evaluated and released to the public for research.
%malignancy
%although the LIDC-IDRI dataset is large, some CT scans were acquired with low dose radiation and reconstructed with large voxel spacing and slice thickness.
%the quality of CT scans is not ensured. 
%correlation study and  boost the performance of nodule malignancy prediction.
% for research
%meticulous
%to facilitate
%, including both handcrafted and deep learning features.
%greatly
%Although radiologists graded the malignancy with two-phase reading, ultimately
%. Handcrafted features and deep learning features may
%quantifying surrounding structures
%, statistical analysis and logistic regression

\section{Conclusion}

We investigated the relationship between pleurae, airways and vessels surrounding a nodule and nodule malignancy on a large public chest CT dataset. Correlation analysis on quantified structures demonstrated that the distance from nodule to pleurae, airways and vessels, together with counting number and normalized volume of airways and vessels, can be viewed as potential lung cancer biomarkers. Features of either arteries or veins benefit nodule diagnosis, which could be useful in further studies on lung cancer.

%makes it clear that their features correlate with nodule malignancy. 
%inside VOI

\section*{Acknowledgments}

This work was partly supported by National Natural Science Foundation of China (Nos. 62003208, 61661010), National Key R\&D Program of China (No. 2019YFB1311503), Committee of Science and Technology, Shanghai, China (Nos. 19411963900, 19510711200), Shanghai Sailing Program (No. 20YF1420800), Shanghai Municipal of Science and Technology (Project No. 20JC1419500), International Research Project METISLAB, Program PHC-Cai Yuanpei 2018 (No. 41400TC).

% and China Scholarship Council (No. 201906230173).

The authors acknowledged the National Cancer Institute and the National Institutes of Health for their critical roles in the creation of the LIDC-IDRI database. The authors also thanked Dr. Yutong Xie (Northwestern Polytechnical University) for the enlightening comments.

%Foundation for the

\section*{Conflicts of Interest}
The authors have no conflicts to disclose.

\section*{References}

\end{document}